\documentclass[11pt,a4paper]{article}
\usepackage{acl2019}
\usepackage{times}
\usepackage{latexsym}
\usepackage{graphicx}
\usepackage{tabularx}
\usepackage{booktabs}
\usepackage{multirow}
\usepackage{arydshln}
\usepackage{caption}

\usepackage{url}
\usepackage{subcaption}
\usepackage{multirow}
\usepackage{makecell}
\usepackage{titlesec}

\usepackage{amsmath,amsfonts,bm}

\def\eqref#1{equation~\ref{#1}}

\def\1{\bm{1}}

\def\rve{{\mathbf{e}}}

\def\rvx{{\mathbf{x}}}

\def\rvz{{\mathbf{z}}}

\DeclareMathAlphabet{\mathsfit}{\encodingdefault}{\sfdefault}{m}{sl}
\SetMathAlphabet{\mathsfit}{bold}{\encodingdefault}{\sfdefault}{bx}{n}

\aclfinalcopy 
\def\aclpaperid{2238} 

\titlespacing{\paragraph}{%
  0pt}{
  0.3\baselineskip}{
  1em}%

\newcommand{\z}{\rvz}
\newcommand{\x}{\rvx}
\newcommand{\e}{\rve}

\title{Cross-Lingual Syntactic Transfer through \\ Unsupervised Adaptation of Invertible Projections}

\author{Junxian He$^1$, Zhisong Zhang$^1$, Taylor Berg-Kirkpatrick$^2$, and Graham Neubig$^1$ \\
  $^1$Language Technologies Institute, Carnegie Mellon University \\
  $^2$Department of Computer Science and Engineering, University of California San Diego \\
  {\small \texttt{\{junxianh,zhisongz,gneubig\}@cs.cmu.edu}, \texttt{tberg@eng.ucsd.edu}}}

\date{}

\begin{document}
\maketitle
\begin{abstract}
Cross-lingual transfer is an effective way to build syntactic analysis tools in low-resource languages. However, transfer is difficult when transferring to typologically distant languages, especially when neither annotated target data nor parallel corpora are available. In this paper, we focus on methods for cross-lingual transfer to distant languages and propose to learn a generative model with a structured prior that utilizes labeled source data and unlabeled target data jointly. The parameters of source model and target model are softly shared through a regularized log likelihood objective. An invertible projection is employed to learn a new interlingual latent embedding space that compensates for imperfect cross-lingual word embedding input.  We evaluate our method on two syntactic tasks: part-of-speech (POS) tagging and dependency parsing. On the Universal Dependency Treebanks, we use English as the only source corpus and transfer to a wide range of target languages. On the 10 languages in this dataset that are distant from English, our method yields an average of 5.2\% absolute improvement on POS tagging and 8.3\% absolute improvement on dependency parsing over a direct transfer method using state-of-the-art discriminative models.\footnote{Code is available at \url{https://github.com/jxhe/cross-lingual-struct-flow}.}
\end{abstract}

\section{Introduction}
\label{sec:intro}
Current top performing systems on syntactic analysis tasks such as part-of-speech (POS) tagging and dependency parsing rely heavily on large-scale annotated data~\citep{huang2015bidirectional,dozat2016deep,ma2018stack}. However, because creating syntactic treebanks is an expensive and time consuming task, annotated data is scarce for many languages. Prior work has demonstrated the efficacy of cross-lingual learning methods \citep{guo2015cross, tiedemann2015cross,guo2016representation,zhang2016ten,ammar2016many,ahmad2018near,schuster2019cross}, which transfer models between different languages through the use of shared features such as cross-lingual word embeddings~\citep{smith2017offline,conneau2017word} or universal part-of-speech tags~\citep{petrov2012universal}.
In the case of \textit{zero-shot} transfer (i.e. with no target-side supervision), a common practice is to train a strong supervised system on the source language and directly apply it to the target language over these shared embedding or POS spaces.
This method has demonstrated promising results, particularly for transfer of models to closely related target languages~\citep{ahmad2018near,schuster2019cross}.

\begin{figure}[!t]
\centering
    \centering
     \begin{subfigure}[b]{0.235\textwidth}
         \centering
         \includegraphics[width=\textwidth]{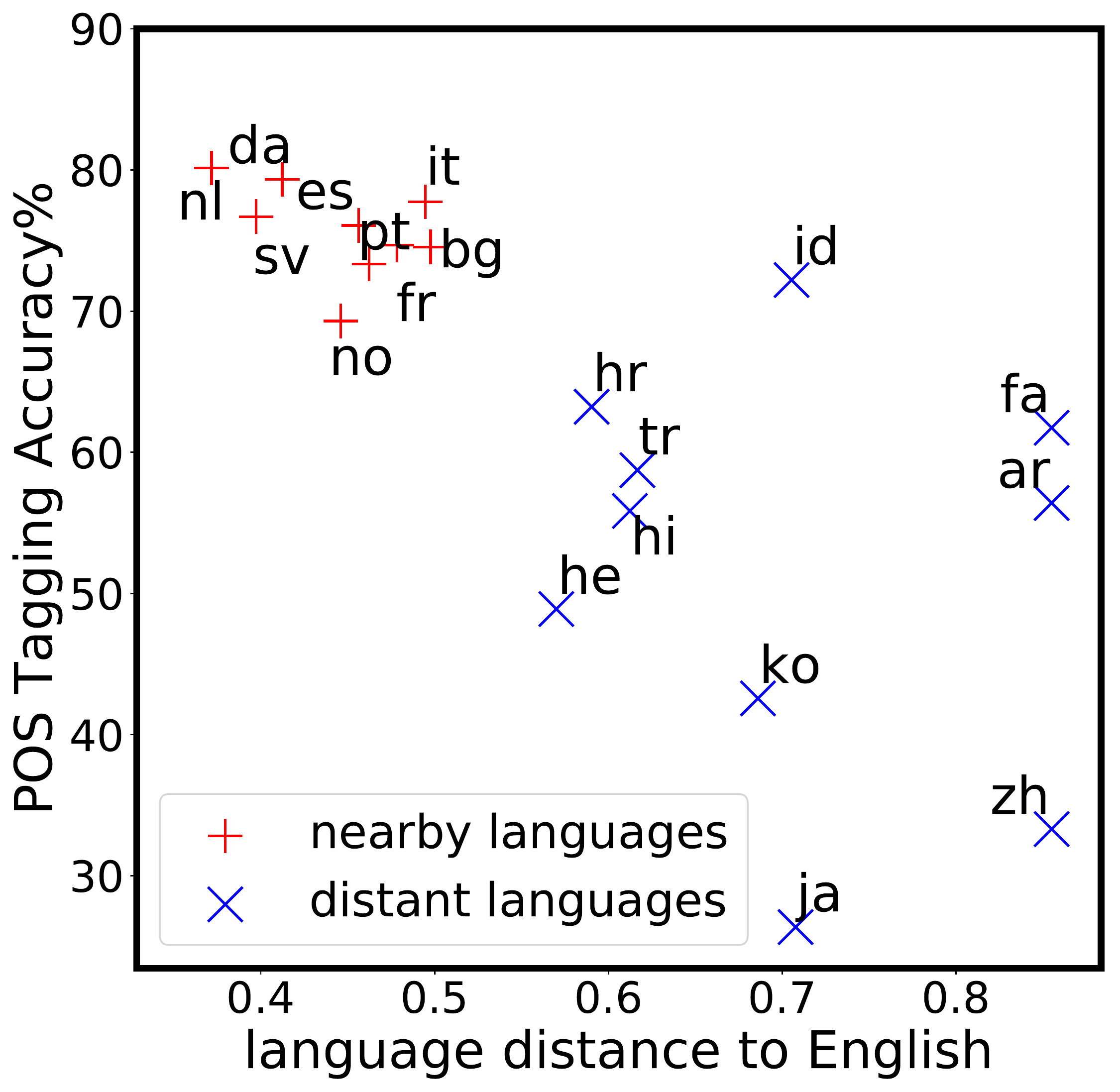}
     \end{subfigure}
     \hfill
     \centering
     \begin{subfigure}[b]{0.235\textwidth}
         \centering
         \includegraphics[width=\textwidth]{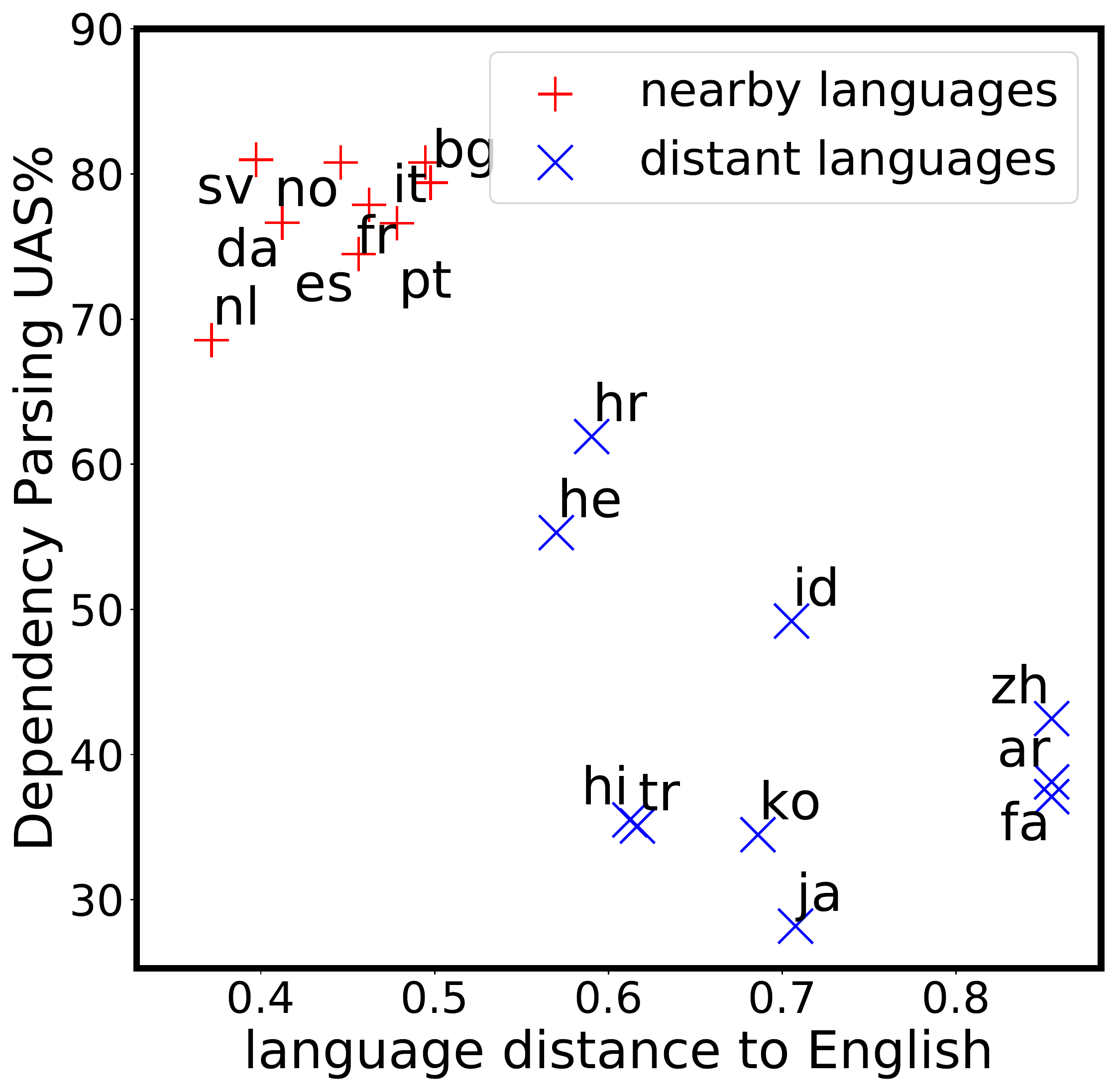}
     \end{subfigure}
\caption{\label{fig:intro} \textbf{Left:} POS tagging transfer accuracy of the Bidirectional LSTM-CRF model, \textbf{Right:} Dependency parsing transfer UAS of the ``SelfAtt-Graph'' model~\citep{ahmad2018near}. These models are trained on the labeled English corpus and directly evaluated on different target languages. The $x$-axis represents language distance to English (details in Section~\ref{sec:distance}). Both models take pre-trained cross-lingual word embeddings as input. The parsing model also uses gold universal POS tags.}
\end{figure}

However, this direct transfer approach often produces poor performance when transferring to more distant languages that are less similar to the source. For example, in Figure~\ref{fig:intro} we show the results of direct transfer of POS taggers and dependency parsers trained on only English and evaluated on 20 target languages using pre-trained cross-lingual word embeddings, where the $x$-axis shows the linguistic distance from English calculated according to the URIEL linguistic database~\citep{littell2017uriel} (more details in Section \ref{sec:difficulties}). As we can see, these systems suffer from a large performance drop when applied to distant languages. The reasons are two-fold: (1) Cross-lingual word embeddings of distant language pairs are often poorly aligned with current methods that make strong assumptions of orthogonality of embedding spaces~\citep{smith2017offline,conneau2017word}. (2) Divergent syntactic characteristics make the model trained on the source language non-ideal, even if the cross-lingual word embeddings are of high quality. 

In this paper we take a drastically different approach from most previous work:
instead of directly transferring a \emph{discriminative} model trained only on labeled data in another language, we use a \emph{generative} model that can be trained in an supervised fashion on labeled data in another language, but also perform unsupervised training to directly maximize likelihood of the target language.
This makes it possible to specifically adapt to the language that we would like to analyze, both with respect to the cross-lingual word embeddings and the syntactic parameters of the model itself.

Specifically, our approach builds on two previous works.
We follow a training strategy similar to \citet{zhang2016ten}, who have previously demonstrated that it is possible to do this sort of cross-lingual unsupervised adaptation, although limited to the sort of linear projections that we argue are too simple for mapping between embeddings in distant languages.
To relax this limitation, we follow \citet{he2018unsupervised} who, in the context of fully unsupervised learning, propose a method using invertible projections (which is also called \textit{flow}) to learn more expressive transformation functions while nonetheless maintaining the ability to train in an unsupervised manner to maximize likelihood.
We learn this structured flow model (detailed in Section~\ref{sec:model}) on both labeled source data and unlabeled target data through a soft parameter sharing scheme.
We describe how to apply this method to two syntactic analysis tasks: POS tagging with a hidden Markov model (HMM) prior and dependency parsing with a dependency model with valence (DMV;~\citet{klein2004corpus}) prior (Section~\ref{sec:exp-dep}).

We evaluate our method on Universal Dependencies Treebanks (v2.2)~\citep{11234/1-2837}, where English is used as the only labeled source data.
10 distant languages and 10 nearby languages are selected as the target without labels. On 10 distant transfer cases -- which we focus on in this paper -- our approach achieves an average of 5.2\% absolute improvement on POS tagging and 8.3\% absolute improvement on dependency parsing over strong discriminative baselines. We also analyze the performance difference between different systems as a function of language distance, and provide preliminary guidance on when to use generative models for cross-lingual transfer.


\section{Difficulties of Cross-Lingual Transfer on Distant Languages}
\label{sec:difficulties}

In this section, we demonstrate the difficulties involved in performing cross-lingual transfer to distant languages.
Specficially, we investigate the direct transfer performance as a function of language distances by training a high-performing system on English and then apply it to target languages.
We first introduce the measurement of language distances and selection of 20 target languages, then study the transfer performance change on POS tagging and dependency parsing tasks. 

\subsection{Language Distance}
\label{sec:distance}
\begin{table}[!t]
    \centering
   \small
    \begin{tabularx}{\columnwidth}{p{1.4cm} | p{5.8cm}}
    \hline
    \textbf{Language Category} & \textbf{Language Names} \\
    \hline
        Distant & Chinese (zh, 0.86), Persian (fa, 0.86), \newline Arabic (ar, 0.86), Japanese (ja, 0.71), \newline Indonesian (id, 0.71), Korean (ko, 0.69), \newline Turkish (tr, 0.62),  Hindi (hi, 0.61), \newline Croatian (hr, 0.59), Hebrew (he, 0.57)\\
    \hline
   Nearby & Bulgarian (bg, 0.50), Italian (it, 0.50), \newline Portuguese (pt, 0.48), French (fr, 0.46), \newline Spanish (es, 0.46), Norwegian (no, 0.45) \newline Danish (da, 0.41), Swedish (sv, 0.40) \newline Dutch (nl, 0.37), German (de, 0.36)\\
\hline
    \end{tabularx}
    \caption{20 selected target languages. Numbers in the parenthesis denote the distances to English.}
    \label{tab:lang-list}
\end{table}

To quantify language distances, we make use of the URIEL~\citep{littell2017uriel} database,\footnote{\url{http://www.cs.cmu.edu/~dmortens/uriel.html}} which represents over 8,000 languages as information-rich typological, phylogenetic, and geographical vectors. These vectors are sourced and predicted from a variety of linguistic resources such as WALS~\citep{wals-81}, PHOIBLE~\citep{phoible}, Ethnologue~\citep{ethnologue}, and Glottolog~\citep{glottolog}. Based on these vectors, this database provides ready-to-use distance statistics between any pair of languages included in the database in terms of various metrics including \textit{genetic} distance, \textit{geographical} distance, \textit{syntactic} distance, \textit{phonological} distance, and phonetic \textit{inventory} distance. These distances are represented by values between 0 and 1. Since phonological and inventory distances mainly characterize intra-word phonetic/phonological features that have less effect on word-level language composition rules, we remove these two and take the average of genetic, geographic, and syntactic distances as our distance measure. 

We rank all languages in Universal Dependencies (UD) Treebanks (v2.2)~\citep{11234/1-2837} according to their distances to English, with the distant ones on the top. Then we select 10 languages from the top that represent the \textit{distant language} group, and 10 languages from the bottom that represent the \textit{nearby language} group. The selected languages are required to meet the following two conditions: (1) at least 1,000 unlabeled training sentences present in the treebank since a reasonably large amount of unlabeled data is needed to study the effect of unsupervised adaptation, and (2) an offline pre-trained word embedding alignment matrix is available.\footnote{Following~\citet{ahmad2018near}, we use the offline pre-trained alignment matrix present in \url{https://github.com/Babylonpartners/fastText_multilingual}, which contains alignment matrices for 78 languages, which also allows comparison with their numbers in Section~\ref{sec:exp-dep}.}
The 20 selected target languages are shown in Table~\ref{tab:lang-list}, which contains distant languages like Persian and Arabic, but also closely related languages like Spanish and French. Detailed statistical information of the selected languages and corresponding treebanks can be found in Appendix~\ref{apd:treebank}.

\subsection{Observations}
\label{sec:obs}
In the direct transfer experiments, we use the pre-trained cross-lingual fastText word embeddings~\citep{bojanowski2017enriching}, aligned with the method of~\citet{smith2017offline}. These embeddings are fixed during training otherwise the alignment would be broken. We employ a bidirectional LSTM-CRF~\citep{huang2015bidirectional} model for POS tagging using NCRF++ toolkit~\citep{yang2018ncrf}, and use the
``SelfAtt-Graph'' model~\citep{ahmad2018near} for dependency parsing.\footnote{We use an implementation and English source model checkpoint identical to the original paper.} Following~\citet{ahmad2018near}, for dependency parsing gold POS tags are also used to learn POS tag embeddings as universal features. We train the systems on English and directly evaluate them on the target languages. Results are shown in Figure~\ref{fig:intro}. While these systems achieve quite accurate results on closely related languages, we observe large performance drops on both tasks as distance to English increases. 
These results motivate our proposed approach, which aims to close this gap by directly adapting to the target language through unsupervised learning over unlabeled text.

\section{Proposed Method}
In this section, we first introduce the unsupervised monolingual models presented in~\citet{he2018unsupervised}, which we refer to as \textit{structured flow models}, then we propose our approach that extends the structured flow models to cross-lingual settings.

\subsection{Unsupervised Training of Structured Flow Models}
\label{sec:model}
\begin{figure}[!t]
\centering
    \includegraphics[scale=0.3]{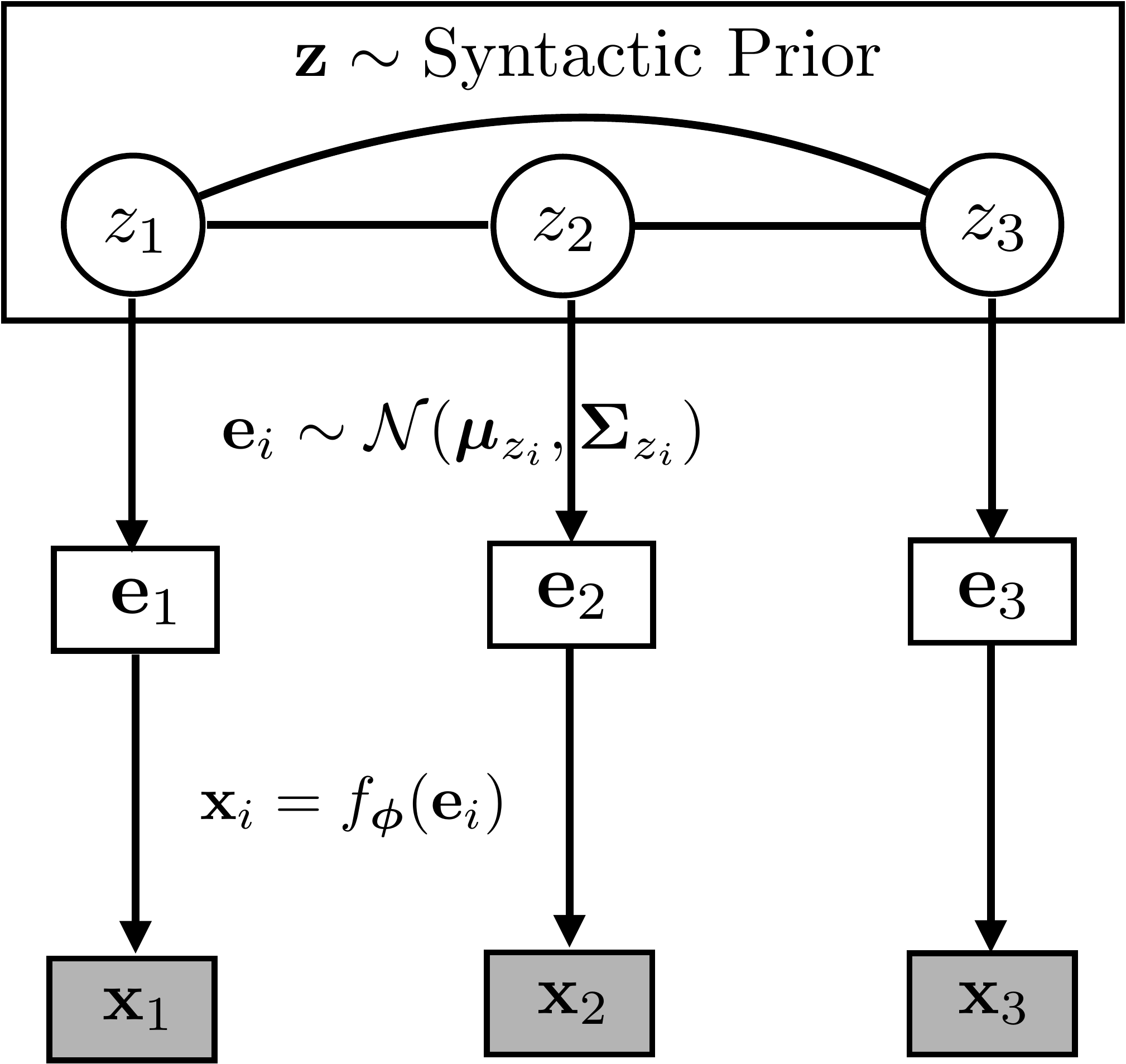}
\caption{Graphical representation of the structured flow model. We denote discrete syntactic variables as $\z$, latent embedding variable as $\e$, and observed pre-trained word embeddings as $\x$. $f_{\bm{\phi}}$ is the invertible projection function.}
\label{fig:model}
\end{figure}

The structured flow generative model, proposed by~\citet{he2018unsupervised}, is a state-of-the-art technique for inducing syntactic structure in a monolingual setting without supervision. This model cascades a structured generative prior $p_{\text{syntax}}(\z; \bm{\theta})$ with an invertible neural network $f_{\bm{\phi}}(\z)$ to generate pre-trained word embeddings $\x = f_{\bm{\phi}}(\z)$, which correspond to the words in the training sentences. $\z$ represents latent syntax variables that are not observed during training. The structured prior defines a probability over syntactic structures, and can be a Markov prior to induce POS tags or DMV prior~\citep{klein2004corpus} to induce dependency structures. Notably, the model side-steps discrete words, and instead uses pre-trained word embeddings as observations, which allows it to be directly employed in cross-lingual transfer setting by using cross-lingual word embeddings as the observations. A graphical illustration of this model is shown in Figure~\ref{fig:model}. Given a sentence of length $l$, we denote $\z=\{z\}_{k=1}^K$ as a set of discrete latent variables from the structured prior, $\e=\{\e_i\}_{i=1}^l$ as the latent embeddings, and $\x=\{\x_i\}_{i=1}^l$ as the observed word embeddings. Note that the number of latent syntax variables $K$ is no smaller than the sentence length $l$, and we assume $\x_i$ is generated (indirectly) conditioned on $\z_i$ for notation simplicity. The model is trained by maximizing the following marginal data likelihood:
\begin{equation}
\label{eq:us-obj}
\begin{split}
p_{\text{us}}(\x) &= \sum\nolimits_{\z}\Big(p_{\text{syntax}}(\z; \bm{\theta}) \\
& \cdot\prod\nolimits_{i=1}^{\ell}p_{\bm{\eta}}(f^{-1}_{\bm{\phi}}(\x_i) | z_i)\Big|\text{det}\frac{\partial f^{-1}_{\bm{\phi}}}{\partial \x_i}\Big|\Big).
\end{split}
\end{equation}
$p_{\bm{\eta}}(\cdot | z_i)$ is defined to be a conditional Gaussian distribution that emits latent embedding $\e$. The projection function $f_{\bm{\phi}}$ projects the latent embedding $\e$ to the observed embedding $\x$. $\frac{\partial f^{-1}_{\bm{\phi}}}{\partial \x_i}$ is the Jacobian matrix of function $f_{\bm{\phi}}^{-1}$ at $\x_i$, and $\big|\text{det}\frac{\partial f^{-1}_{\bm{\phi}}}{\partial \x_i}\big|$ represents the absolute value of its determinant. 

To understand the intuitions behind Eq.~\ref{eq:us-obj}, first denote the log likelihood over the latent embedding $\e$ as $\log p_{\text{gaus}}(\cdot)$, then log of Eq.~\ref{eq:us-obj} can be equivalently rewritten as:
\begin{equation}
\label{eq:us-obj-s}
\begin{split}
\log p_{\text{us}}(\x) &= \log p_{\text{gaus}}(f_{\bm{\phi}}^{-1}(\x)) \\
& + \sum\nolimits_{i=1}^l\log \Big|\text{det}\frac{\partial f^{-1}_{\bm{\phi}}}{\partial \x_i}\Big|.
\end{split}
\end{equation}
Eq.~\ref{eq:us-obj-s} shows that $f_{\phi}^{-1}(\x)$ inversely projects $\x$ to a new latent embedding space, on which the unsupervised training objective is simply the Gaussian log likelihood with an additional Jacobian regularization term. The Jacobian regularization term accounts for the volume expansion or contraction behavior of the projection, thus maximizing it can be thought of as preventing information loss.\footnote{Maximizing the Jacobian term encourages volume expansion and prevents the latent embedding from collapsing to a (nearly) single point.} This projection scheme can flexibly transform embedding space to fit the task at hand, but still avoids trivial solutions by preserving information. 

While $f_{\bm{\phi}}^{-1}(\x)$ can be any invertible function,~\citet{he2018unsupervised} use a version of the NICE architecture~\citep{dinh2014nice} to construct $f_{\bm{\phi}}^{-1}$, which has the advantage that the determinant term is constantly equal to one. This structured flow model allows for exact marginal data likelihood computation and exact inference by the use of dynamic programs to marginalize out $\z$. More details about this model can be found in~\citet{he2018unsupervised}. 

\subsection{Supervised Training of Structured Flow Models}
While~\citet{he2018unsupervised} train the structured flow model in an unsupervised fashion, this model can be also trained with supervised data when $\z$ is observed. Supervised training is required in the cross-lingual transfer where we train a model on the high-resource source language. The supervised objective can be written as: 

\begin{equation}
\label{eq:s-obj}
\begin{split}
p_{\text{s}}(\z, \x) &= p_{\text{syntax}}(\z; \bm{\theta}) \\
& \cdot\prod\nolimits_{i=1}^{\ell}p_{\bm{\eta}}(f^{-1}_{\bm{\phi}}(\x_i) | z_i)\Big|\text{det}\frac{\partial f^{-1}_{\bm{\phi}}}{\partial \x_i}\Big|,
\end{split}
\end{equation}

\subsection{Multilingual Training through Parameter Sharing}
In this paper, we focus on the zero-shot cross-lingual transfer setting where the syntactic structure $\z$ is observed for the source language but unavailable for the target languages. Eq.~\ref{eq:us-obj-s} is an unsupervised objective which is optimized on the target languages, and Eq.~\ref{eq:s-obj} is optimized on the source language.
To establish connections between the source and target languages, we employ two instances of the structured flow model -- a \textit{source model} and a \textit{target model} -- and share parameters between them. The source model uses the supervised objective, Eq.~\ref{eq:s-obj}, and the target model uses the unsupervised objective, Eq.~\ref{eq:us-obj-s}, and both are optimized jointly. Instead of tying their parameters in a hard way, we share their parameters softly through an L2 regularizer that encourages similarity. We use subscript $p$ to represent variables of the source model and $q$ to represent variables of the target model. Together, our joint training objective is:
\begin{equation}
\label{eq:adapt-obj}
\begin{split}
\hspace{-5pt}
&L(\bm{\theta}_{\{p, q\}}, \bm{\eta}_{\{p, q\}}, \bm{\phi}_{\{p,q\}})\hspace{-2pt} = \hspace{-2pt}\log p_{\text{s}}(\x_p) \hspace{-2pt}+\hspace{-2pt} \log p_{\text{us}}(\x_q) \\
&- \frac{\beta_1}{2}\|\bm{\theta}_p - \bm{\theta}_q\|^2 - \frac{\beta_2}{2}\|\bm{\eta}_p - \bm{\eta}_q\|^2 \\
&- \frac{\beta_3}{2}\|\bm{\phi}_p - \bm{\phi}_q\|^2,
\end{split}
\raisetag{1.1\baselineskip}
\end{equation}  
where $\bm{\beta}=\{\beta_1, \beta_2, \beta_3\}$ are regularization parameters. Introduction of hyperparameters is concerning because in the zero-shot transfer setting we do not have annotated data to select the parameters for each target language, but in experiments we found it unnecessary to tune $\bm{\beta}$ for different target languages separately, and it is possible to use the same $\bm{\beta}$ within the same language category (i.e.\ distant or nearby). Under the parameter sharing scheme the projected latent embedding space $\e$ can be understood as the new interlingual embedding space from which we learn the syntactic structures. The expressivity of the flow model used in learning this latent embeddings space is expected to compensate for the imperfect orthogonality between the two embedding spaces.

Further, jointly training both models with Eq.~\ref{eq:adapt-obj} is more expensive than typical cross-lingual transfer setups -- it would require re-training both models for each language pair. To improve efficiency and memory utilization, in practice we use a simple pipelined approach: (1) We pre-train parameters for the source model only once, in isolation. (2) We use these parameters to initialize each target model, and regularize all target parameters towards this initializer via the L2 terms in Eq.~\ref{eq:adapt-obj}. 
In this way, we only need to save the pre-trained parameters for a single source model, and target-side fine-tuning converges much faster than training each pair from scratch. This training approximation has been used before in~\citet{zhang2016ten}.  

\section{Experiments}
In this section, we first describe the dataset and experimental setup, and then report the cross-lingual transfer results of POS tagging and dependency parsing on distant target languages. Lastly we include analysis of different systems.   

\subsection{Experimental Setup}
Across both POS tagging and dependency parsing tasks, we run experiments on Universal Dependency Treebanks (v2.2)~\citep{11234/1-2837}. Specifically, we train the proposed model on the English corpus with annotated data and fine-tune it on target languages in an unsupervised way. In the rest of the paper we will use \textit{Flow-FT} to term our proposed method. We use the aligned cross-lingual word embeddings described in Section~\ref{sec:obs} as the observations of our model. To compare with~\citet{ahmad2018near}, on dependency parsing task we also use universal gold POS tags to index tag embeddings as part of observations. Specifically, the tag embeddings are concatenated with word embeddings to form $\x$, tag embeddings are updated when training on the source language, and fixed at fine-tuning stage. 
We implement the structured flow model based on the public code from~\citet{he2018unsupervised},\footnote{\url{https://github.com/jxhe/struct-learning-with-flow}.}which contains models with Markov prior for POS tagging and DMV prior for dependency parsing. Detailed hyperparameters can be found in Appendix~\ref{apd:hyperparam}. Both source model and target model are optimized with Adam~\citep{kingma2014adam}. Training on the English source corpus is run 5 times with different random restarts for all models, then the source model with the best English test accuracy is selected to perform transfer.

We compare our method with a direct transfer approach that is based on the state-of-the-art discriminative models as described in Section~\ref{sec:obs}. The pre-trained cross-lingual word embeddings for all models are frozen since fine-tuning them will break the multi-lingual alignments. In addition, to demonstrate the efficacy of unsupervised adaptation, we also include direct transfer results of our model without fine-tuning, which we denote as \textit{Flow-Fix}. On the POS tagging task we re-implement the generative baseline in~\citet{zhang2016ten} that employs a linear projection (\textit{Linear-FT}). We present results on 20 target languages in ``distant languages'' and ``nearby languages'' categories to analyze the difference of the systems and the scenarios to which they are applicable.


\subsection{Part-Of-Speech Tagging}
\label{sec:exp-tag}
\paragraph{Setup. }Our method aims to predict coarse universal POS tags, as fine-grained tags are language-dependent. The discriminative baseline with the NCRF++ toolkit~\citep{yang2018ncrf} achieves supervised test accuracy on English of 94.02\%, which is competitive (rank 12) on the CoNLL 2018 Shared Task scoreboard that uses the same dataset.\footnote{For reference, check the ``en\_ewt'' treebank results in http://universaldependencies.org/conll18/results-upos.html.} The regularization parameters $\bm{\beta}$ in all generative models are tuned on the Arabic\footnote{We choose Arabic simply because it is first in alphabetical order.} development data and kept the same for all target languages. Our running $\bm{\beta}$ is $\beta_1=0, \beta_2=500, \beta_3=80$. Unsupervised fine-tuning is run for 10 epochs. 

\begin{table}[!t]
    \centering
    \hspace{-0.25cm}
    \resizebox{1. \columnwidth}{!}{
    \begin{tabularx}{1.2\columnwidth}{lcrrr}
    \toprule
    & \multicolumn{1}{c}{Discriminative} & \multicolumn{3}{c}{Generative} \\
    \textbf{\small Lang} & \textbf{\small LSTM-CRF} & \textbf{\small Flow-Fix} & \textbf{\small Flow-FT} & \textbf{\small Linear-FT}\\
    \hline
    \multicolumn{5}{c}{Distant Languages} \vspace{0.7mm}\\
        zh (0.86) & 33.31 & 35.24 & \textbf{43.44} & 35.95\\
        fa (0.86) & 61.74 & 55.32 & \textbf{64.47} & 34.35\\
        ar (0.86) & 56.41 & 49.70 & \textbf{64.00} & 38.95 \\
        ja (0.71) & 26.37 & 25.09 & \textbf{38.37} & 12.49 \\
        id (0.71) & 72.21 & 63.73 & \textbf{73.51} & 57.56\\ 
        ko (0.69) & \textbf{42.57} & 39.56 & 41.76 & 18.30 \\
        tr (0.62) & 58.74 & 43.17 & \textbf{60.08} & 22.79\\
        hi (0.61) & 55.85 & 47.18 & \textbf{64.75} & 38.04\\
        hr (0.59) & \textbf{63.23} & 50.57 & 57.90 & 56.53\\
        he (0.57) & 48.90 & 47.97 & \textbf{62.69} & 48.17\\ \hdashline
        AVG & 51.93 & 45.75 & \textbf{57.10} & 36.31\\
    \midrule
    \multicolumn{5}{c}{Nearby Languages} \vspace{0.7mm}\\
        bg (0.50) & \textbf{74.55} & 62.18 & 64.69 & 66.71\\
        it (0.50) & 77.75 & 69.93 & \textbf{80.99} & 73.55 \\
        pt (0.48) & \textbf{74.68} &65.08 & 72.65  & 72.54 \\
        fr (0.46) & \textbf{73.33} & 64.15 & 69.78 & 66.63\\
        es (0.46) & 76.07 & 65.77 & \textbf{77.19} & 72.86 \\
        no (0.45) & \textbf{69.30} & 58.98 & 62.05 & 62.38\\
        da (0.41) & \textbf{79.33} & 62.42 & 68.68 & 67.31\\
        sv (0.40) & \textbf{76.70} & 58.91 & 66.34 & 61.82\\
        nl (0.37) & \textbf{80.15} & 66.52 & 68.74 & 66.08\\
        de (0.36) & \textbf{68.75} & 57.91 & 59.97 & 56.16\\ \hdashline
        AVG & \textbf{75.06} & 63.19 & 69.11 & 66.60\\ \hline
        en$^{\ast}$ & 94.02 & 87.03 & -- & 84.69 \\
    \bottomrule 
    \end{tabularx}}
    \caption{POS tagging accuracy results (\%). Numbers next to languages names are their distances to English. Supervised accuracy on English ($\ast$) is included for reference.}
    \label{tab:res-tag}
\end{table}

\paragraph{Results. }We show our results in Table~\ref{tab:res-tag}, where unsupervised fine-tuning achieves considerable and consistent performance improvements over the Flow-Fix baseline in both language categories. When compared the discriminative LSTM-CRF baseline, our approach outperforms it on 8 out of 10 distant languages, with an average of 5.2\% absolute improvement. Unsurprisingly, however, it also underperforms the expressive LSTM-CRF on 8 out of 10 nearby languages. The reasons for this phenomenon are two-fold. First, the flexible  LSTM-CRF model is better able to fit the source English corpus than our method (94.02\% vs 87.03\% accuracy), thus it is also capable of fitting similar input when transferring. Second, unsupervised adaptation helps less when transferring to nearby languages (5.9\% improvement over Flow-Fix versus 11.3\% on distant languages), we posit that this is because a large portion of linguistic knowledge is shared between similar languages, and the cross-lingual word embeddings have better quality in this case, so unsupervised adaptation becomes less necessary. While the Linear-FT baseline on nearby languages is comparable to our method, its performance on distant languages is much worse, which confirms the importance of invertible projection, especially when language typologies are divergent.

\subsection{Dependency Parsing}
\label{sec:exp-dep}
\paragraph{Setup. }In preliminary parsing results we found that transferring to ``nearby language'' group is likely to suffer from catastrophic forgetting~\citep{mccloskey1989catastrophic} and thus requires stronger regularization towards the source model. This also makes sense intuitively since nearby languages should prefer the source model more than distant languages.  Therefore, we use two different sets of regularization parameters for nearby languages and distant languages, respectively. Specifically, $\bm{\beta}$ for the ``distant languages'' group is set as $\beta_1=\beta_2=\beta_3=0.1$, tuned on the Arabic development set, and for the ``nearby languages'' group $\bm{\beta}$ is set as $\beta_1=\beta_2=\beta_3=1$, tuned on the Spanish development set. Unsupervised adaptation is performed on sentences of length less than 40 due to memory constraints,\footnote{Reducing batch size can address this memory issue, but greatly increases the training time.} but we test on sentences of all lengths. We run unsupervised fine-tuning for 5 epochs, and evaluate using unlabeled attachment score (UAS) with punctuation excluded. 

\paragraph{Results. }We show our results in Table~\ref{tab:res-parse}. While unsupervised fine-tuning improves the performance on the distant languages, it only has minimal effect on nearby languages, which is consistent with our observations in the POS tagging experiment and implies that unsupervised adaption helps more for distant transfer. Similar to POS tagging results, our method is able to outperform state-of-the-art ``SelfAtt-Graph'' model on 8 out of 10 distant languages, with an average of 8.3\% absolute improvement, but the strong discriminative baseline performs better when transferring to nearby languages. Note that the supervised performance of our method on English is poor. This is mainly because the DMV prior is too simple and limits the capacity of the model. While this model still achieves good performance on distant transfer, incorporating more complex DMV variants~\citep{jiang2016unsupervised} might lead to further improvement.

\begin{table}[!t]
    \centering
    \resizebox{1. \columnwidth}{!}{
    \begin{tabularx}{1.2\columnwidth}{lcc r r}
    \toprule
    & \qquad\qquad & \multicolumn{1}{c}{Discriminative} & \multicolumn{2}{c}{Generative} \\
    \textbf{\small Lang} && \textbf{\small SelfAtt-Graph} & \textbf{\small Flow-Fix} & \textbf{\small Flow-FT} \\
    \hline
    \multicolumn{5}{c}{Distant Languages} \vspace{0.7mm}\\
        zh (0.86) && \textbf{42.48} & 35.72 & 37.26 \\
        fa (0.86) && 37.10 & 37.58 & \textbf{63.20} \\
        ar (0.86) && 38.12 & 32.14 & \textbf{55.44} \\
        ja (0.71) && 28.18 & 19.03 & \textbf{43.75} \\
        id (0.71) && 49.20 & 46.74 & \textbf{64.20} \\ 
        ko (0.69) && 34.48 & 34.76 & \textbf{37.03} \\
        tr (0.62)  && 35.08 & 34.76 & \textbf{36.05} \\
        hi (0.61)  && \textbf{35.50}& 29.20 & 33.17 \\
        hr (0.59) && 61.91& 59.57 &\textbf{65.31} \\
        he (0.57) && 55.29 & 51.35 & \textbf{64.80} \\ \hdashline
        AVG && 41.73 & 38.09 & \textbf{50.02} \\
    \midrule
    \multicolumn{5}{c}{Nearby Languages} \vspace{0.7mm}\\
        bg (0.50) && \textbf{79.40} & 73.52 & 73.57 \\
        it (0.50) && \textbf{80.80} & 68.84 & 70.68 \\
        pt (0.48) && \textbf{76.61}& 66.61 & 66.61 \\
        fr (0.46) && \textbf{77.87}& 65.92 & 67.66  \\
        es (0.46) && \textbf{74.49} & 63.10 & 64.28 \\
        no (0.45) && \textbf{80.80}& 65.48 & 65.29 \\
        da (0.41) && \textbf{76.64}& 61.64 & 61.08 \\
        sv (0.40) && \textbf{80.98}& 66.22 & 64.43 \\
        nl (0.37) && \textbf{68.55}& 61.59 &61.72 \\
        de (0.36) && \textbf{71.34}& 70.10 & 69.52 \\ \hdashline
        AVG && \textbf{76.75}& 66.30 & 66.48 \\ \midrule
        en$^{\ast}$ && 91.82 & 67.80 & -- \\
    \bottomrule 
    \end{tabularx}}
    \caption{Dependency parsing UAS (\%) on sentences of all lengths. Numbers next to languages names are their distances to English. Supervised accuracy on English ($\ast$) is included for reference.}
    \label{tab:res-parse}
\end{table}

\paragraph{Analysis on Dependency Relations. }
\begin{figure*}[t]
	\centering
	\begin{subfigure}{0.329\textwidth}
		\centering
		\includegraphics[width=\textwidth]{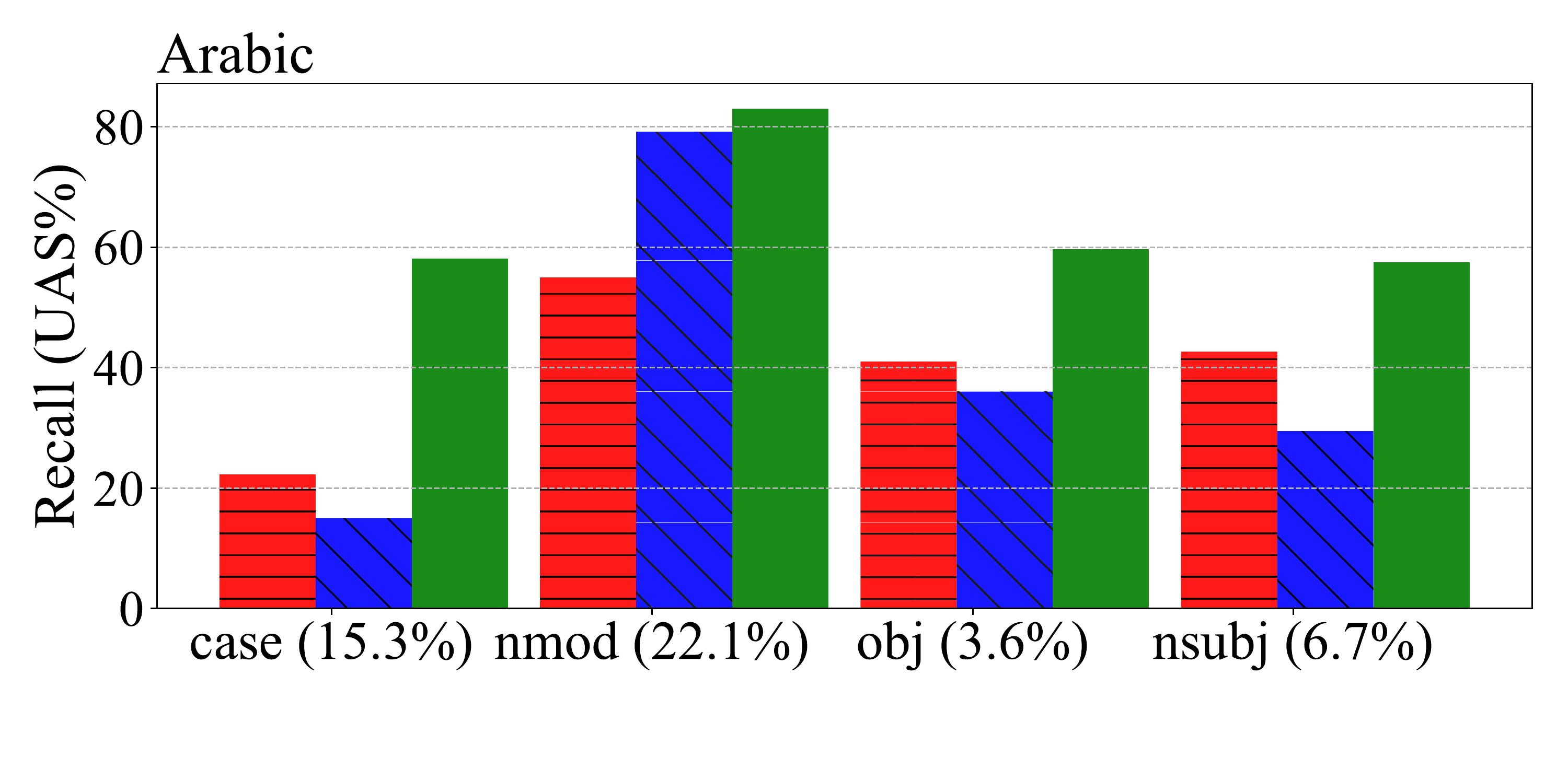}
	\end{subfigure}
	\hfill
	\begin{subfigure}{0.329\textwidth}
		\centering
		\includegraphics[width=\textwidth]{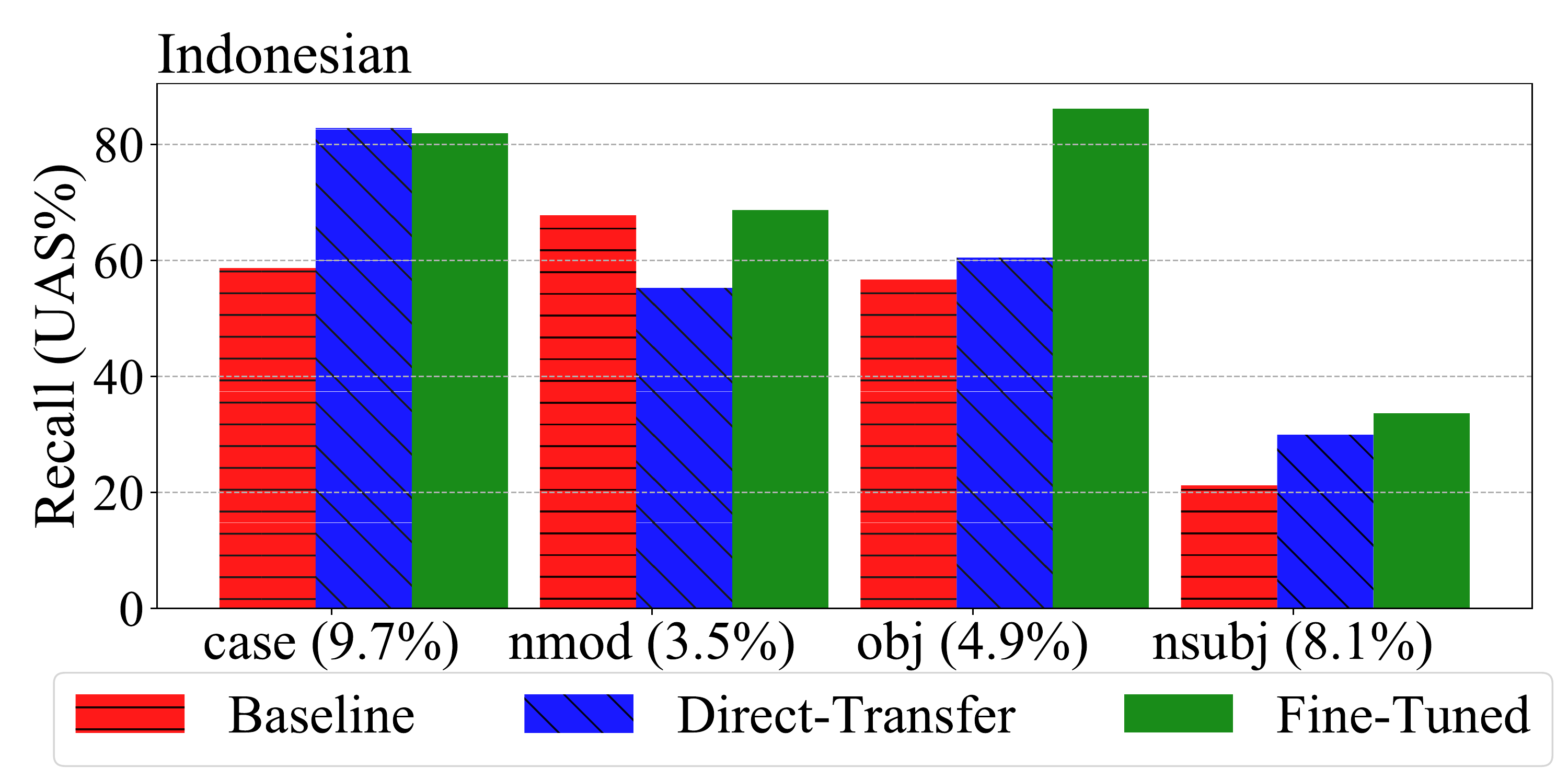}
	\end{subfigure}
	\hfill
	\begin{subfigure}{0.329\textwidth}
		\centering
		\includegraphics[width=\textwidth]{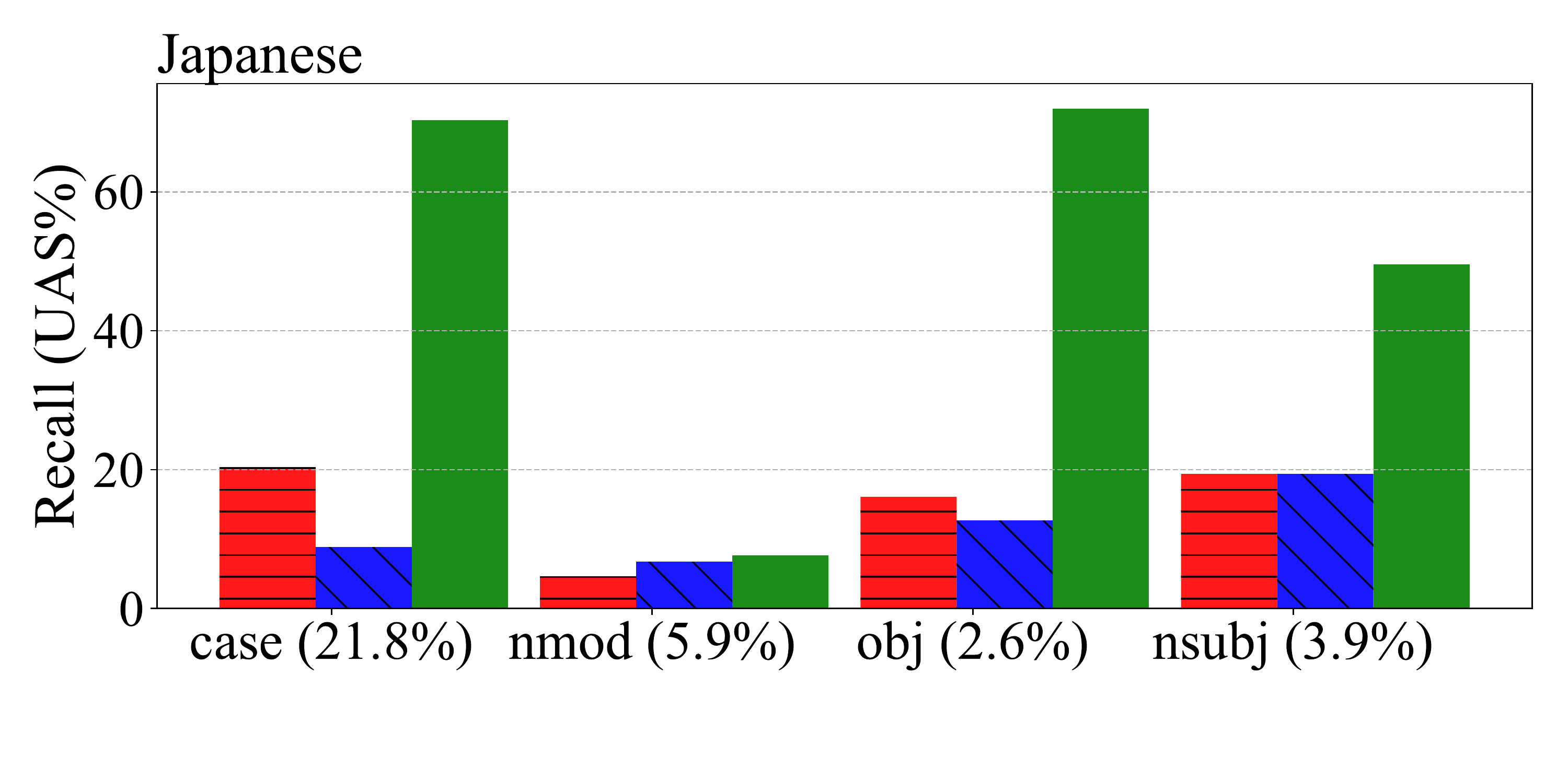}
	\end{subfigure}
	\caption{\label{fig:analysis} Results (UAS\%) on typical dependency relations for Arabic, Indonesian and Japanese, respectively. ``Baseline'' denotes the ``SelfAtt-Graph'' model, and ``Direct-Transfer'' denotes our source model without fine-tuning. The number in the parenthesis after each dependency label indicates the relative frequency of this type. }
    \vspace{-5mm}
\end{figure*}

We further perform breakdown analysis on dependency relations to see how unsupervised adaptation helps learn new dependency rules. We select three typical distant languages with different word order of Subject, Object and Verb \cite{wals-81}: Arabic (Modern Standard, VSO), Indonesian (SVO) and Japanese (SOV).

We investigate the unlabeled accuracy (recall) on the gold dependency labels. We especially explore four typical dependency relations: case (case marking), nmod (nominal modifier), obj (object) and nsubj (nominal subject). The first two are ``nominal dependents'' (modifiers for nouns) and the rest two are the main nominal ``core arguments'' (arguments for the predicate). Although different languages may vary, these four types are representative relations and occupies 25\% to 40\% in frequencies among all 37 UD dependency types.

We compare our fine-tuning model with the baseline ``SelfAtt-Graph'' model and our basic model without fine-tuning. As shown in Figure \ref{fig:analysis}, although our direct transfer model obtain similar results when compared with the baseline, the fine-tuning method brings large improvements on most of these dependency relations. In these three languages, Japanese benefits from our tuning method the most, probably because its word order is quite different from English and the baseline may overfit to the English order. For example, in Japanese, almost all of the ``case'' relations are head-first and ``obj'' relations are modifier-first, and these patterns are exactly opposite to those in English, which serves as our source language. As a result, direct transfer models fail on most of these relations since they only learn the patterns in English.
With our fine-tuning on unlabeled data, the model may get more familiar with the unusual patterns of word order and predict more correct attachment decisions (around 0.4 improvements in recalls). 
In Arabic and Indonesian, although not as obviously as in Japanese, the improvements are still consistent, especially on the relations of the core arguments.



\subsection{When to Use Generative Models?}
\label{sec:analysis}
In unsupervised cross-lingual transfer setting, it is hard to find a system that is able to achieve state-of-the-art on all languages. 
As reflected by our experiments, there is a tradeoff between fitting source language and generalizing to target language -- the flexibility of discriminative models results in overfitting issue and poor performance when transferred to distant languages. Unfortunately, a limited number of high-resource languages and many more low-resource languages in the world are mostly distant. This means that distant transfer is a practical challenge we face when dealing with low-resource languages. Next we try to give a preliminary guidance about which system should be used in specific transfer scenarios.

As discussed in Section~\ref{sec:distance}, there are different types of distance metrics. Here we aim to compute the significance of correlation between the performance difference between our method and the discriminative baseline and different distance features. 
We have five input distance features: geographic, genetic, syntactic, inventory, and phonological. 

Specifically, we fit a generalized linear model (GLM) on the difference in accuracy and five features of all 20 target languages, then we perform a hypothesis test to compute the $p$-value that reflects the significance of specific features.\footnote{We use the GLM toolkit present in the H2O Python Module.} 
Results are shown in Table~\ref{tab:p-value}, where we can conclude that the genetic distance feature is significantly correlated with POS tagging performance, while geographic distance feature is significantly correlated with dependency parsing performance. As assumed before, inventory and phonological distance do not have much influence on the transfer. Interestingly, syntactic distance is not the significant term for both tasks, we posit that this is because the transfer performance is affected by both cross-lingual word embedding quality and linguistic features, thus genetic/geographic distance might be a better indicator overall. The results suggest that our method might be more suitable than the discriminative approach at genetically distant transfer for POS tagging and geographically distant transfer for parsing.   

\begin{table}[!t]
\centering
 \resizebox{1.0 \columnwidth}{!}{
\begin{tabular}{lrr}
\toprule
\multicolumn{1}{l}{Feature} &\multicolumn{2}{c}{$p$-value}  \\
\multicolumn{1}{l}{} &\multicolumn{1}{c}{POS tagging} &\multicolumn{1}{c}{Dependency Parsing}
\\ \hline
Geographic  & 0.465 & 0.013 \\
Genetic     & 0.007  & 0.531\\
Syntactic   & 0.716 & 0.231\\
Inventory & 0.982 & 0.453\\
Phonological & 0.502 & 0.669\\
\bottomrule
\end{tabular}}
\caption{$p$-value of different distance features on POS tagging and dependency parsing task. A lower $p$-value indicates stronger association between the feature and the response, which is the difference between our method and the discriminative baselines.}
\label{tab:p-value}
\end{table}

\subsection{Effect of Multilingual-BERT}
\label{sec:bert}
So far the analysis and experiments of this paper focus on non-contextualized fastText word embeddings. We note that concurrently to this work,~\citet{wu2019beto} found that the recently released multilingual BERT (mBERT;~\citet{devlin2018bert}) is able to achieve impressive performance on various cross-lingual transfer tasks. To study the effect of contextualized mBERT word embeddings on our proposed method, we report the average POS tagging and dependency parsing results in Table~\ref{tab:bert-short}, while detailed numbers on each language are included in Appendix~\ref{apd:bert}. In the mBERT experiments, all the settings and hyperparameters are the same as in Section~\ref{sec:exp-tag} and Section~\ref{sec:exp-dep}, but the aligned fastText embeddings are replaced with the mBERT embeddings.\footnote{We use the multilingual cased BERT base model released in \url{https://github.com/google-research/bert}.} We also include the average results from fastText embeddings for comparison. 

On the POS tagging task all the models greatly benefit from the mBERT embeddings, especially our method on nearby languages where the mBERT outperforms the fastText by an average of 16 absolute points. Moreover, unsupervised adaptation still considerably improves the Flow-Fix baseline, and surpasses the LSTM-CRF baseline on 9 out of 10 distant languages with an average of 6\% absolute performance boost. Different from the fastText setting where our method underperforms the discriminative baseline on the nearby language group, by the use of mBERT embeddings our method also beats the discriminative baseline on 7 out of 10 nearby languages with an average of 3\% absolute improvement. A major limitation of our method lies in its strong independence assumptions, which results in the failure to model the long-term context information. We posit that the contextualized word embeddings like mBERT exactly compensate for this drawback in our model through incorporating the context information into the observed word embeddings, so that our method is able to outperform the discriminative baseline on both distant and nearby language groups.

\begin{table}[!t]
\small
    \centering
    \begin{tabular}{lrrrr}
    \toprule
    & \multicolumn{2}{c}{Tagging} & \multicolumn{2}{c}{Parsing}\\
    \textbf{emb} & \textbf{Disc} & \textbf{Flow-FT} & \textbf{Disc} & \textbf{Flow-FT} \\
    \hline
    & \multicolumn{4}{c}{Distant Languages} \vspace{0.7mm}\\
        fastText & 51.93 & 57.10 &41.73 &50.02 \\
        mBERT & 60.24 & \bf 66.56 &\bf 51.86 &50.11\\
    \midrule
    & \multicolumn{4}{c}{Nearby Languages} \vspace{0.7mm}\\
        fastText & 75.06 & 69.11 &76.75 & 66.48\\
        mBERT & 82.17 & \bf 85.48 &\bf 83.41 & 67.70\\ 
    \bottomrule 
    \end{tabular}
    \caption{Average of POS tagging accuracy (\%) and dependency parsing UAS (\%) results, comparing mBERT and fastText. ``Disc'' denotes the discriminative baselines.}
    \label{tab:bert-short}
\end{table} 

On dependency parsing task, however, our method does not demonstrate significant improvement by the use of mBERT, while mBERT greatly helps the discriminative baseline. Therefore, although our method still outperforms the discriminative baseline on four very distant languages, the baseline demonstrates superior performance on other languages when using mBERT. Interestingly, we find that the performance of flow-based models with mBERT is similar to the performance with fastText word embeddings.
Based on this, better generative models for unsupervised dependency parsing that can take advantage of contextualized embeddings seems a promising direction for future work.


\section{Related Work}
Cross-lingual transfer learning has been widely studied to help induce syntactic structures in low-resource languages~\citep{mcdonald2011multi,tackstrom2013token,agic2014cross,tiedemann2015cross,kim2017cross,schuster2019cross,ahmad2018near}. In the case when no available target annotations are available, unsupervised cross-lingual transfer can be performed by directly applying pre-trained source model to the target language.~\citep{guo2015cross,schuster2019cross,ahmad2018near}. 
The challenge of direct transfer method lies in the different linguistic rules between source and distant target languages. Utilizing multiple sources of resources can mitigate this issue and has been actively studied in the past years~\citep{cohen2011unsupervised,naseem2012selective,tackstrom2013target,zhang2015hierarchical,aufrant2015zero,ammar2016many,wang2018synthetic,wang2019surface}. Other approaches that try to overcome the lack of annotations include annotation projection by the use of bitext supervision or bilingual lexicons~\citep{hwa2005bootstrapping,smith2009parser,wisniewski2014cross} and source data point selection~\citep{sogaard2011data,tackstrom2013target}. 

Learning from both labeled source data and unlabeled target data has been explored before. ~\citet{cohen2011unsupervised} learns a generative target language parser as a linear interpolation of multiple source language parameters, ~\citet{naseem2012selective} and~\citet{tackstrom2013target} rely on additional language typological features to guide selective model parameter sharing in a multi-source transfer setting, ~\citet{wang2018synthetic,wang2019surface} extract linguistic features from target languages by training a feature extractor on multiple source languages. 


\section{Conclusion}
In this work, we focus on transfer to distant languages for POS tagging and dependency parsing, and propose to learn a structured flow model in a cross-lingual setting. Through learning a new latent embedding space as well as language-specific knowledge with unlabeled target data, our method proves effective at transferring to distant languages.

\section*{Acknowledgements}

This research was supported by NSF Award No. 1761548 ``Discovering and Demonstrating Linguistic Features for Language Documentation'',  and, in part, by an Amazon Research Award to the third author.

\bibliography{acl2019}
\bibliographystyle{acl_natbib}

\newpage
\clearpage
\appendix

\section{Details of UD Treebanks}
\label{apd:treebank}
\begin{table}[!h]
\centering
	\small
	\resizebox{0.9 \columnwidth}{!}{
	\begin{tabular}{l|c|c|c c}
		\hline
		\hline
		Language & Dist. & Treebank & & \#Sent. \\
		\hline
		\multirow{3}{*}{Chinese (zh)} & \multirow{3}{*}{0.86} & \multirow{3}{*}{GSD} & train & 3997 \\
		& & & dev & 500 \\
		& & & test & 500 \\
		\hline
		\multirow{3}{*}{Persian (fa)} & \multirow{3}{*}{0.86} & \multirow{3}{*}{Seraji} & train & 4798 \\
		& & & dev & 599 \\
		& & & test & 600 \\
		\hline
		\multirow{3}{*}{Arabic (ar)} & \multirow{3}{*}{0.86} & \multirow{3}{*}{PADT} & train & 6075 \\
		& & & dev & 909 \\
		& & & test & 680 \\
		\hline
		\multirow{3}{*}{Japanese (ja)} & \multirow{3}{*}{0.71} & \multirow{3}{*}{GSD} & train & 7164 \\
		& & & dev & 511 \\
		& & & test & 557 \\
		\hline
		\multirow{3}{*}{Indonesian (id)} & \multirow{3}{*}{0.71} & \multirow{3}{*}{GSD} & train & 4477 \\
		& & & dev & 559 \\
		& & & test & 557 \\
		\hline
		\multirow{3}{*}{Korean (ko)} & \multirow{3}{*}{0.69} & \multirow{3}{*}{\makecell{GSD,\\Kaist}} & train & 27410 \\
		& & & dev & 3016 \\
		& & & test & 3276 \\
		\hline
		\multirow{3}{*}{Turkish (tr)} & \multirow{3}{*}{0.62} & \multirow{3}{*}{IMST} & train & 3685 \\
		& & & dev & 975 \\
		& & & test & 975 \\
		\hline
		\multirow{3}{*}{Hindi (hi)} & \multirow{3}{*}{0.61} & \multirow{3}{*}{HDTB} & train & 13304 \\
		& & & dev & 1659 \\
		& & & test & 1684 \\
		\hline
		\multirow{3}{*}{Croatian (hr)} & \multirow{3}{*}{0.59} & \multirow{3}{*}{SET} & train & 6983 \\
		& & & dev & 849 \\
		& & & test & 1057 \\
		\hline
		\multirow{3}{*}{Hebrew (he)} & \multirow{3}{*}{0.57} & \multirow{3}{*}{HTB} & train & 5241 \\
		& & & dev & 484 \\
		& & & test & 491 \\
		\hline
		\multirow{3}{*}{Bulgarian (bg)} & \multirow{3}{*}{0.50} & \multirow{3}{*}{BTB} & train & 8907 \\
		& & & dev & 1115 \\
		& & & test & 1116 \\
		\hline
		\multirow{3}{*}{Italian (it)} & \multirow{3}{*}{0.50} & \multirow{3}{*}{ISDT} & train & 13121 \\
		& & & dev & 564 \\
		& & & test & 482 \\
		\hline
		\multirow{3}{*}{Portuguese (pt)} & \multirow{3}{*}{0.48} & \multirow{3}{*}{\makecell{Bosque,\\GSD}} & train & 17993 \\
		& & & dev & 1770 \\
		& & & test & 1681 \\
		\hline
		\multirow{3}{*}{French (fr)} & \multirow{3}{*}{0.46} & \multirow{3}{*}{GSD} & train & 14554 \\
		& & & dev & 1478 \\
		& & & test & 416 \\
		\hline
		\multirow{3}{*}{Spanish (es)} & \multirow{3}{*}{0.46} & \multirow{3}{*}{\makecell{GSD,\\AnCora}} & train & 28492 \\
		& & & dev & 3054 \\
		& & & test & 2147 \\
		\hline
		\multirow{3}{*}{Norwegian (no)} & \multirow{3}{*}{0.45} & \multirow{3}{*}{\makecell{Bokmaal,\\Nynorsk}} & train & 29870 \\
		& & & dev & 4300 \\
		& & & test & 3450 \\
		\hline
		\multirow{3}{*}{Danish (da)} & \multirow{3}{*}{0.41} & \multirow{3}{*}{DDT} & train & 4383\\
		& & & dev & 564 \\
		& & & test & 565 \\
		\hline
		\multirow{3}{*}{Swedish (sv)} & \multirow{3}{*}{0.40} & \multirow{3}{*}{Talbanken} & train & 4303 \\
		& & & dev & 504 \\
		& & & test & 1219 \\
		\hline
		\multirow{3}{*}{Dutch (nl)} & \multirow{3}{*}{0.37} & \multirow{3}{*}{\makecell{Alpino,\\LassySmall}} & train & 18058\\
		& & & dev & 1394 \\
		& & & test & 1472 \\
		\hline
		\multirow{3}{*}{German (de)} & \multirow{3}{*}{0.36} & \multirow{3}{*}{GSD} & train & 13814 \\
		& & & dev & 799 \\
		& & & test & 977 \\
		\hline
		\multirow{3}{*}{English (en)} & \multirow{3}{*}{--} & \multirow{3}{*}{EWT} & train & 12543 \\
		& & & dev & 2002 \\
		& & & test & 2077 \\
		\hline
		\hline
	\end{tabular}}
    \caption{Statistics of the UD Treebanks that we used.}
\end{table}

We list the statistics of the UD Treebanks that we used in the following two tables. The left one lists the distance (to English) languages and the right one lists the similar (to English) languages.

\section{Model Hyperparameters}
\label{apd:hyperparam}
We use the same architecture as in~\citet{he2018unsupervised} for the invertible projection function $f_{\bm{\phi}}$ which is the NICE architecture~\citep{dinh2014nice}. It contains 8 coupling layers. The coupling function in each coupling layer is a rectified network with an input layer, one hidden layer, and linear output units. The number of hidden units is set to the same as the number of input units, which is 150 in our case. POS tagger is trained with batch size 32, while dependency parser is trained with batch size 16.

\section{Full Results with mBERT}
\label{apd:bert}
Here we report in Table~\ref{tab:bert} the full results on all languages with mBERT.\footnote{The results of our discriminative baselines are different from the ones reported in~\citet{wu2019beto} because they do not use additional encoders on top of the pretrained mBERT word embeddings, while we keep the models unchanged here for direct comparison with fastText embeddings. On some languages our version produces better results and sometimes their version is superior.}
\begin{table*}[!t]
\small
    \centering
    \begin{tabular}{lcrrrrrr}
    \toprule
    & &\multicolumn{3}{c}{POS Tagging} & \multicolumn{3}{c}{Dependency Parsing}\\
    \textbf{Lang} &\qquad\qquad\qquad\qquad & \textbf{LSTM-CRF} & \textbf{Flow-Fix} & \textbf{Flow-FT} & \textbf{SelfAtt-Graph} & \textbf{Flow-Fix} & \textbf{Flow-FT} \\
    \hline
    & \multicolumn{7}{c}{Distant Languages} \vspace{0.7mm}\\
        zh (0.86) && 59.63 & 53.61 & \bf 65.84 &\bf 48.78 &35.73 &35.64 \\
        fa (0.86) && 57.63 & 56.18 & \bf 68.55 &51.47 &37.99 &\bf 63.18\\
        ar (0.86) && 53.50 & 48.92 & \bf 67.33 &50.91 &32.13 &\bf 56.85\\
        ja (0.71) && \bf 46.81 & 40.98 & 46.06 &40.08 &19.23 &\bf 43.55\\
        id (0.71) && 74.95 & 70.95 & \bf 78.72 &57.94 &47.00 &\bf 64.35\\ 
        ko (0.69) && 50.74 & 47.99 & \bf 54.07 &\bf 39.42 &34.67 &37.02\\
        tr (0.62)  && 60.08 & 54.69 & \bf 61.16 &\bf 42.80 &34.88 & 37.06\\
        hi (0.61)  && 58.86& 53.16 & \bf 68.39 &\bf 48.44 &29.15 &33.17\\
        hr (0.59) && 74.98& 66.35 &\bf 78.61 &\bf 73.63 &59.68 &65.27\\
        he (0.57) && 65.24 & 57.27 & \bf 76.83 &\bf 65.11 &51.39 & 65.03\\ \hdashline
        AVG (mBERT) && 60.24 & 55.01 & \bf 66.56 &\bf 51.86 &38.19 &50.11\\
        AVG (fastText) && 51.93 & 45.75 & 57.10 &41.73 &38.09 &50.02 \\
    \midrule
    & \multicolumn{7}{c}{Nearby Languages} \vspace{0.7mm}\\
    \hline
        bg (0.50) && \bf 82.36 & 74.56 & 80.68 &\bf 86.32 &73.65&74.06\\
        it (0.50) && 76.70 & 66.02 & \bf 87.88 &\bf 86.71& 69.09&71.59\\
        pt (0.48) && 83.45& 80.83 & \bf 86.49 &\bf 83.75&66.67&69.56\\
        fr (0.46) && 79.22 & 74.21 & \bf 87.21 &\bf 86.64& 66.08& 69.14\\
        es (0.46) && 77.68 & 72.28 & \bf 84.50& \bf 81.74& 63.18&66.46\\
        no (0.45) && \bf 85.29& 80.69 & 83.96 &\bf 85.01 & 65.47& 66.08\\
        da (0.41) && 85.57& 81.90 & \bf 86.79 &\bf 82.22 &61.61 &62.15\\
        sv (0.41) && \bf 86.39& 81.27 & 86.31 &\bf 85.33 & 66.04& 64.51\\
        nl (0.40) && 83.67& 78.88 & \bf 85.05 & \bf 77.32& 61.70&63.24\\
        de (0.37) && 81.37& 78.97 &\bf 85.96 &\bf 79.03 & 70.19& 70.19\\\hdashline
        AVG (mBERT) && 82.17& 76.96 & \bf 85.48 &\bf 83.41& 66.37& 67.70\\ 
        AVG (fastText) && 75.06 & 63.19 & 69.11 &76.75 &66.30 & 66.48\\
        \midrule
        en$^{\ast}$ && 95.13 & 91.22 & -- &92.84 &67.76&--\\
    \bottomrule 
    \end{tabular}
    \caption{POS tagging accuracy (\%) and dependency parsing UAS (\%) results when using mBERT as the aligned embeddings. Numbers next to languages names are their distances to English. Supervised accuracy on English ($\ast$) is included for reference.}
    \label{tab:bert}
\end{table*}

\end{document}